%% file: emnlp2023.tex
\pdfoutput=1

\documentclass[11pt]{article}

\usepackage[]{EMNLP2023}

\usepackage{times}
\usepackage{latexsym}

\usepackage[T1]{fontenc}

\usepackage[utf8]{inputenc}

\usepackage{microtype}

\usepackage{inconsolata}

\usepackage{times}
\usepackage{latexsym}
\usepackage{graphicx}
\usepackage{caption}
\usepackage{subcaption}
\usepackage{latexsym}
\usepackage{lscape}
\usepackage{amsmath}
\usepackage{booktabs}
\usepackage{tabularx}
\usepackage{paralist}
\usepackage{amssymb,amsmath,amsthm,enumitem}

\usepackage{amssymb}
\usepackage[]{xcolor}
\usepackage{etoolbox}
\usepackage{multirow}
\usepackage{soul}
\usepackage{hyperref}
\usepackage{amsmath}
\usepackage{latexsym}

\newcommand{\spartqa}{\textsc{SpartQA}}
\newcommand{\resq}{\textsc{ReSQ}}
\newcommand{\spartun}{\textsc{SpaRTUN}}
\newcommand{\pistaq}{\textsc{PistaQ}}
\newcommand{\etor}{\textsc{PistaQ}}
\newcommand{\rtoe}{\textsc{SREQA}}

\newcommand{\sreqa}{\textsc{SREQA}}

\newcommand{\babi}{\text{bAbI}}
\newcommand{\bert}{\text{BERT}}
\newcommand{\berteq}{\text{BERT-EQ}}
\newcommand{\bertspartun}{\text{BERT*}}

\newcommand{\qtype}{\textsc{Q-Type}}
\newcommand{\msprl}{\textsc{mSpRL}}
\newcommand{\sprl}{\textsc{SpRL}}
\newcommand{\auto}{\textsc{SpartQA-Auto}}
\newcommand{\human}{\textsc{SpartQA-Human}}

\title{Disentangling Extraction and Reasoning
in  Multi-hop Spatial Reasoning}

\author{Roshanak Mirzaee \\
  Michigan State university \\
  \texttt{mirzaeem@msu.edu} \\\And
  Parisa Kordjamshidi \\
  Michigan State university \\
  \texttt{kordjams@msu.edu} \\}

\begin{document}
\maketitle
\begin{abstract}
Spatial reasoning over text is challenging as the models not only need to extract the direct spatial information from the text but also reason over those and infer implicit spatial relations. Recent studies highlight the struggles even large language models encounter when it comes to performing spatial reasoning over text.
In this paper, we explore the potential benefits of disentangling the processes of information extraction and reasoning in models to address this challenge.
To explore this, we design various models that disentangle extraction and reasoning~(either symbolic or neural) and compare them with state-of-the-art~(SOTA) baselines with no explicit design for these parts.
Our experimental results consistently demonstrate the efficacy of disentangling, showcasing its ability to enhance models' generalizability within realistic data domains. 
\end{abstract}

\input{01-intro}

\input{02-related}
\input{03-pistaq}
\input{04-experiments}

\input{05-results}
\input{10-conclusion}

\section*{Acknowledgements}
This project is partially supported by the National Science Foundation~(NSF) CAREER award 202826. Any opinions,
findings, and conclusions or recommendations expressed in this material are those of the authors and
do not necessarily reflect the views of the National
Science Foundation. We thank all reviewers for their helpful comments and suggestions. We would like to express our gratitude to Hossein Rajaby Faghihi for his valuable discussions and input to this paper.

\section*{Limitations}
Our model is evaluated on a Spatial Reasoning task using specifically designed spatial logical rules. However, this methodology can be readily extended to other reasoning tasks that involve a limited set of logical rules, which can be implemented using logic programming techniques.
The extraction modules provided in this paper are task-specific and do not perform well on other domains, but they can be fine-tuned on other tasks easily. Using LLM in the extraction phase can also deal with this issue. Also, using \msprl{} annotation on \resq{}(which this data is provided on) decreases the performance of our models. This annotation does not contain the whole existing relations in the context. The evaluation of the reasoning module is based on the existing datasets. However, we cannot guarantee that they cover all possible combinations between spatial rules and relation types. Many questions in \resq{} need spatial commonsense to be answered. As a result, due to the limitation of our symbolic spatial reasoner, the performance of the pipeline model is much lower than what we expected. Due to the high cost of training GPT3.5 on large synthetic data, we cannot fine-tune the whole GPT3.5 and only provide the GPT3.5 with $Few\_shot$ learning on small human-generated benchmarks. Also, due to the limited access, we can only test PaLM2 and GPT4 on a few examples.

\bibliography{anthology,custom}
\bibliographystyle{acl_natbib}

\newpage
\appendix

\input{99-appendix}




\end{document}

%% file: 01-intro.tex
\section{Introduction}
\label{sec:introduction}

Despite the high performance of recent pretrained language models on question-answering~(QA) tasks, solving questions that require multi-hop reasoning is still challenging~\cite{mavi2022survey}.
In this paper, we focus on spatial reasoning over text which can be described as inferring the implicit\footnote{By implicit, we mean indirect relations, not metaphoric usages or implicit meaning for the relations.} spatial relations from explicit relations\footnote{relationships between objects and entities in the environment, such as location, distance, and relative position.} described in the text. 
Spatial reasoning plays a crucial role in diverse domains, including language grounding~\cite{liu-etal-2022-things}, navigation~\cite{zhang-etal-2021-towards}, and human-robot interaction~\cite{venkatesh2021spatial}.
By studying this task, we can analyze both the reading comprehension and logical reasoning capabilities of models.



\begin{figure}[t]
    \centering
    \includegraphics[width=\linewidth]{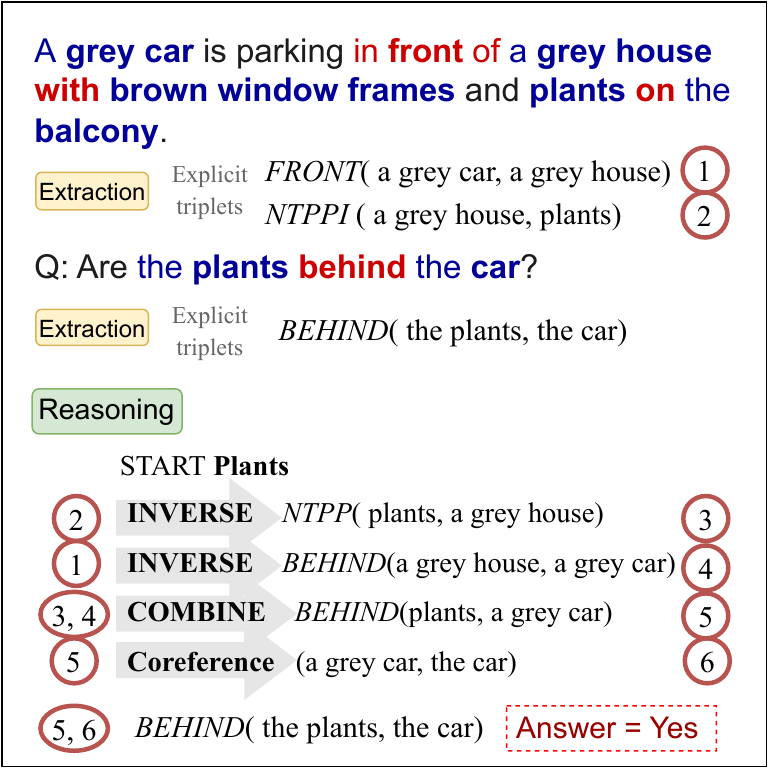}
    \caption{An example of steps of spatial reasoning on \resq{} dataset. 
    We begin by searching for \textit{the plants} from the question triplet within the text, enabling us to extract explicit triplets~(1,2). Next, we apply rules such as INVERSE to deduce implicit triplets~(3,4,5). Then, utilizing triplets 5 and 6 we determine the final answer, 'Yes'. 
    NTPP: Non-Tangential Proper Part~(Table~\ref{tab:spatial_relations}).}
    \label{fig:resq}
\end{figure}





Previous work has investigated the use of general end-to-end deep neural models such as pretrained language models~(PLM)~\cite{mirzaee-etal-2021-spartqa} in spatial question answering~(SQA).
PLMs show reasonable performance on the SQA problem and can implicitly learn spatial rules from a large set of training examples. However, 
the black-box nature of PLMs makes it unclear whether these models are making the abstractions necessary for spatial reasoning or their decisions are based solely on patterns observed in the data. 

As a solution for better multi-hop reasoning, recent research has investigated the impact of using fine-grained information extraction modules such as Named Entity Recognition~(NER)~\cite{molla2006named,mendes2010named}, gated Entity/Relation ~\cite{zheng2021relational} 
or semantic role labels~(SRL)~\cite{shen-lapata-2007-using,faghihi2023role} on the performance of models.

On a different thread, cognitive studies~\cite{stenning2012human, dietz2015computational} show when the given information is shorter, humans also find spatial abstraction and use spatial rules to infer implicit information. Figure~\ref{fig:resq} shows an example of such extractions.
Building upon these findings, we aim to address the limitations of end-to-end models and capitalize on the advantages of fine-grained information extraction in solving SQA.
Thus, we propose models which disentangle the\textit{ language understanding} and \textit{spatial reasoning} computations as two separate components. 
Specifically, we first design a pipeline model that includes trained neural modules for extracting direct fine-grained spatial information from the text and performing symbolic spatial reasoning over them.

\begin{table}[t]
\small

\resizebox{\linewidth}{!}{%
\begin{tabular}{|l|l|l|}
\hline
\textbf{\begin{tabular}[c]{@{}l@{}}Formalism\\ (General Type)\end{tabular}} & \textbf{Spatial Type} & \textbf{Expressions (e.g.)} \\ \hline
\begin{tabular}{l}
    Topological  \\
     (RCC8)
\end{tabular} & \begin{tabular}[c]{@{}l@{}}DC (disconnected)\\ EC (Externally Connected)\\ PO (Partially Overlapped)\\ EQ (Equal)\\ TPP\\ NTPP\\ TPPI\\ NTPPI \end{tabular} & \begin{tabular}[c]{@{}l@{}}disjoint\\ touching\\ overlapped\\ equal\\ covered by\\ in, inside\\ covers\\ has\end{tabular} \\ \hline
\begin{tabular}{l}
     Directional \\
     (Relative)
\end{tabular} & \begin{tabular}[c]{@{}l@{}}LEFT, RIGHT\\ BELOW, ABOVE\\ BEHIND, FRONT\end{tabular} & \begin{tabular}[c]{@{}l@{}}left of, right of\\ under, over\\ behind, in front\end{tabular} \\ \hline
\begin{tabular}{l}
     Distance  
\end{tabular} & Far, Near & far, close \\ 
\hline
\end{tabular}%
}
\caption{List of spatial relation formalism and types.}
\label{tab:spatial_relations}
\end{table}

The second model is simply an end-to-end PLM  that uses annotations used in extraction modules of pipeline model in the format of \textit{extra QA} supervision. This model aims to demonstrate the advantages of using separate extraction modules compared to a QA-based approach while utilizing  the same amount of supervision. 
Ultimately, the third model is an end-to-end PLM-based model on relation extraction tasks that has explicit latent layers to disentangle the extraction and reasoning inside the model. This model incorporates a neural spatial reasoner, which is trained to identify all spatial relations between each pair of entities.

We evaluate the proposed models on multiple SQA datasets, demonstrating the effectiveness of the disentangling extraction and reasoning approach in controlled and realistic environments. 
Our pipeline outperforms existing SOTA models by a significant margin on benchmarks with a controlled environment~(toy tasks) while utilizing the same or fewer training data. 
However, in real-world scenarios with higher ambiguity of natural language for extraction and more rules to cover, our end-to-end model with explicit layers for extraction and reasoning performs better.



These results show that disentangling extraction and reasoning benefits deterministic spatial reasoning and improves generalization in realistic domains despite the coverage limitations and sensitivity to noises in symbolic reasoning. 
These findings highlight the potential of leveraging language models for information extraction tasks and emphasize the importance of explicit reasoning modules rather than solely depending on black-box neural models for reasoning.

%% file: 02-related.tex
\section{Related Research}
\label{sec:background}

\noindent \textbf{End-to-end model on SQA:} To solve SQA tasks, recent research evaluates the performance of different deep neural models such as Memory networks~\cite{shi2022stepgame,sukhbaatar2015end}, Self-attentive Associative Memory~\cite{le2020self}, subsymbolic fully connected neural network~\cite{zhu2022reasoning}, and Recurrent Relational Network (RRN)~\cite{palm2017recurrent}. 
\citeauthor{spartun,mirzaee-etal-2021-spartqa} use transfer learning and provide large synthetic supervision that enhances the performance of PLMs on spatial question answering.
However, the results show a large gap between models and human performance on human-generated data.
Besides, none of these models use explicit spatial semantics to solve the task.
The only attempt towards integrating spatial semantics into spatial QA task is a baseline model introduced in~\cite{mirzaee-etal-2021-spartqa}, which uses rule-based spatial semantics extraction for reasoning on \babi{}~(task 17) which achieves 100\% accuracy without using any training data.


\noindent \textbf{Extraction and Reasoning:}  
While prior research has extensively explored the use of end-to-end models for learning the reasoning rules~\cite{minervini2020learning, qu2020rnnlogic}, there is limited discussion on separating the extraction and reasoning tasks. 
\citeauthor{nye2021dualsystem} utilizes LMs to generate new sentences and extract facts while using some symbolic rules to ensure consistency between generated sentences.
Similarly, ThinkSum~\cite{ozturkler2022thinksum} uses LMs for knowledge extraction~(Think) and separate probabilistic reasoning~(Sum), which sums the probabilities of the extracted information. However, none of these works are on multi-step or spatial Reasoning.



%% file: 03-pistaq.tex
\section{Proposed Models}
\label{sec:proposed_model}

To understand the effectiveness of disentangling the extraction and reasoning modules, we provide three groups of models.
The first model is a pipeline of extraction and symbolic reasoning~(\S\ref{sec:pistaq}), the second model is an end-to-end PLM that uses the same spatial information supervision but in a QA format~(\S\ref{sec:bert-eq}), and the third model is an end-to-end neural model with explicit layers of extraction and reasoning~(\S\ref{sec:sreqa}). We elaborate  each of these models in the subsequent sections.


\paragraph{Task} The target task is spatial question answering (SQA), which assesses models' ability to comprehend spatial language and reason over it. Each example includes a textual story describing entities and their spatial relations, along with questions asking an \textit{implicit} relation between entities~(e.g., Figure~\ref{fig:resq}). 
SQA benchmarks provide two types of questions: YN~(Yes/No) queries about the existence of a relation between two groups of entities, and FR~(Find Relation) seeks to identify all possible~(direct/indirect) relations between them. The answer to these questions is chosen from a provided candidate list. For instance, the candidate list for FR questions can be a sublist of all relation types in Table~\ref{tab:spatial_relations}.


\subsection{Pipeline of Extraction and Reasoning}
\label{sec:pistaq}
\begin{figure}[t]
    \centering
    \includegraphics[width = \linewidth]{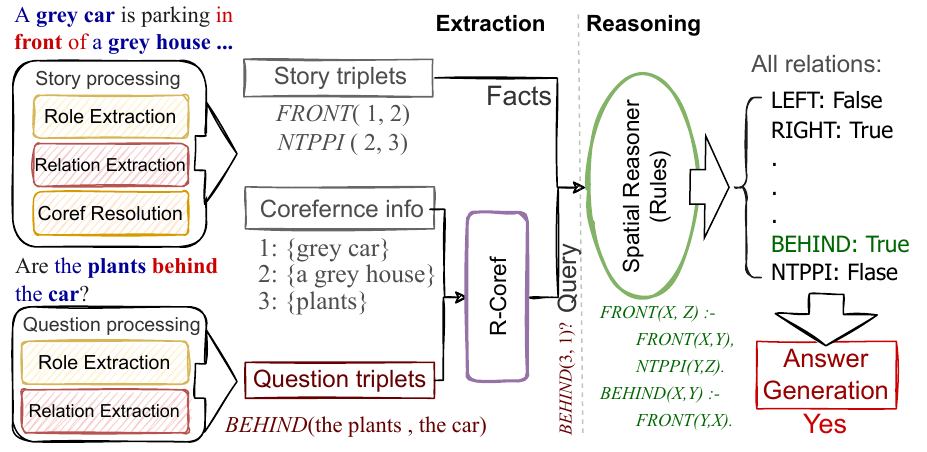}
    \caption{\etor{} pipeline based on disentangled extraction and reasoning. 
    In this model, facts, e.g., FRONT(grey car, grey house), are extracted from the story and linked by coreference modules. The R-Coref equates `the car' from the question with `a grey car' in the story and forms a query. This query, along with facts, is sent to the spatial reasoner. Finally, the spatial reasoner employs FRONT and BEHIND rules and returns True as the answer.}
    \label{fig:pistaq-etor}
    \vspace{-5mm}
\end{figure}

Here, we describe our suggested pipeline model designed for spatial question answering task, referred to as \textbf{\etor{}}\footnote{\textbf{PI}peline model for \textbf{S}pa\textbf{T}i\textbf{A}l \textbf{Q}uestion answering}.  
As shown in the extraction part of Figure~\ref{fig:pistaq-etor}, 
the spatial information is extracted first and forms a set of triplets for a story~(Facts) and a question~(Query). Then a coreference resolution module is used to connect these triplets to each other.
Given the facts and queries, the spatial reasoner infers all implicit relations. The answer generator next conducts the final answer.
Below we describe each module in more detail. 

\begin{table*}[t]

\resizebox{\textwidth}{!}{%

\begin{tabular}{|llllll|}
\hline
&&&&& \\
Not          & $\forall (X,Y) \in Entities$     & $R \in \left \{ Dir \vee PP \right \}$         &  & $\operatorname{IF} R(X,Y)$                       & $\Rightarrow \operatorname{NOT}(R\_reverse(X,Y))$                  \\
Inverse      & $\forall (X,Y) \in Entities$     & $R \in \left \{ Dir \vee PP \right \}$ &  & $\operatorname{IF} R(Y,X)$                       & $\Rightarrow R\_reverse(X,Y)$                                      \\
Symmetry     & $\forall (X,Y) \in Entities$     & $R \in \left \{ Dis \vee (RCC - PP) \right \}$ &  & $\operatorname{IF} R(Y,X)$                       & $\Rightarrow R(X,Y)$                                               \\
Transitivity & $\forall (X,Y, Z) \in Entities$  & $R \in \left \{ Dir \vee PP \right \}$         &  & $\operatorname{IF} R(X,Z), R(Z,Y) $              & \begin{tabular}[c]{@{}l@{}}$\Rightarrow  R(X,Y)$\end{tabular} \\
Combination  & $\forall (X,Y,Z,H) \in Entities$ & $R \in Dir, *PP \in PP $                       &  & $\operatorname{IF} *PP(X,Z), R(Z,H), *PPi(Z,Y) $ & \begin{tabular}[c]{@{}l@{}}$\Rightarrow  R(X,Y)$ 
\end{tabular}\\
&&&&&\\
\hline
\end{tabular}%
}
\caption{Designed spatial rules~\cite{spartun}.  $Dir$: Directional relations~(e.g., LEFT), $Dis$: Distance relations~(e.g., FAR), $PP$: all Proper parts relations~(NTPP, NTPPI, TPPI, TPP), $RCC-PP$: All RCC8 relation except proper parts relations. $*PP$: one of TPP or NTPP. $*PPi$: one of NTPPi or TPPi.}
\label{fig:sp-rules}
\end{table*}  






\noindent \textbf{Spatial Role Labeling~(\sprl{})} is the task of identifying and classifying the \textit{spatial roles} of phrases within a text~(including the Trajector, Landmark, and Spatial Indicator) and formalizing their \textit{relations}~\cite{kordjamshidi2010spatial}.
Here, we use the same \sprl{} modules as in~\cite{spartun}.
This model first computes the token representation of a story and its question using a \bert{} model. 
Then a BIO tagging layer is applied on the tokens representations using~(O, B-entity, I-entity, B-indicator, and I-indicator) tags. Finally, a softmax layer on the BIO tagger output selects the spatial entities\footnote{Trajector/Landmark}~(e.g., `grey car' or `plants' in Figure~\ref{fig:pistaq-etor}) and spatial indicators~(e.g., `in front of' in Figure~\ref{fig:pistaq-etor}).

Given the output of the spatial role extraction module, for each combination of $(\operatorname{Trajector},\ \operatorname{Spatial\ indicator},\ \operatorname{Landmark})$ in each sentence, we create a textual input\footnote{$[\text{CLS}, traj, \text{SEP}, indic, \text{SEP}, land, \text{SEP}, sentence, \text{SEP}]$ } and pass it to a \bert{} model. 
To indicate the position of each spatial role in the sentence, we use segment embeddings and add $1$ if it is a role position and $0$ otherwise.
The $[CLS]$ output of \bert{} will be passed to a one-layer MLP that provides the probability for each triplet.
To apply the logical rules on the triplets, we need to assign a relation type to each triplet.
To this aim, we use another multi-classification layer on the same $[CLS]$ token to identify the spatial types of the triplet. 
The classes are relation types in Table~\ref{tab:spatial_relations} alongside a class $\operatorname{NaN}$ for triplet with no spatial meaning. For instance, in Figure~\ref{fig:pistaq-etor}, $(\operatorname{grey\ car}, \operatorname{in\ front\ of}, \operatorname{grey\ house})$ is a triplet with $FRONT$ as its relation type while $(\operatorname{grey\ house} , \operatorname{in\ front\ of}, \operatorname{grey\ car})$ is not a triplet and its relation type is $NaN$.
We use a joint loss function for triplet and relation type classification to train the model.

\noindent \textbf{Coreference Resolution}
Linking the extracted triplets from the stories is another important step required in this task, as different phrases or pronouns may refer to same entity. To make such connections, we implement a coreference resolution model based on~\cite{lee-etal-2017-end} and extract all antecedents for each entity and assign a unique $id$ to them. 
In contrast to previous work, we have extended the model to support plural antecedents~(e.g., two circles). More details about this model can be found in Appendix~\ref{appendix:coref}.
To find the mentions of the question entities in the story and create the queries, we use a Rule-based Coreference~(R-Coref) based on exact/partial matching. In Figure~\ref{fig:pistaq-etor}, `the car' in the question has the same id as `the grey car' from the story's triplets. 

\noindent\textbf{{Logic-based Spatial Reasoner}}
To do symbolic spatial reasoning, we use the reasoner from~\cite{spartun}. This reasoner is implemented in Prolog and utilizes a set of rules on various relation types, as illustrated in Table~\ref{fig:sp-rules}. 
Given the facts and queries in Prolog format, the spatial reasoner can carry out the reasoning process and provide an answer to any given query. The reasoner matches variables in the program with concrete values and a backtracking search to explore different possibilities for each rule until a solution is found. As shown in Figure~\ref{fig:pistaq-etor}, the reasoner uses a FRONT and a BEHIND rules over the facts and generates the True response for the query.

\subsection{PLMs Using \sprl{} Annotations}
\label{sec:bert-eq}
To have a fair comparison between the QA baselines and models trained on \sprl{} supervision, we design \textbf{\berteq{}}\footnote{\bert{}+\textbf{E}xtra \textbf{Q}uestion}.
We convert the \sprl{} annotation into extra YN questions\footnote{This augmentation does not apply to FR type since it inquires about all relations between the two asked entities.} asking about explicit relations between a pair of entities.
To generate extra questions, we replace triplets from the \sprl{} annotation into the ``Is [Trajector] [Relation*] [Landmark]?'' template. The [Trajector] and [Landmark] are the entity phrases in the main sentence ignoring pronouns and general names~(e.g., ``an object/shape''). The [Relation*] is a relation expression~(examples presented in Table~\ref{tab:spatial_relations}) for the triplet relation type. To have equal positive and negative questions, we reverse the relation in half of the questions. 
We train \berteq{} using both original and extra questions by passing the ``question$+$story'' into a \bert{} with answers classification layers.

\subsection{PLMs with Explicit Extractions}
\label{sec:sreqa}

\begin{figure}[t]
    \begin{subfigure}{\linewidth}%
    \includegraphics[width=\linewidth]{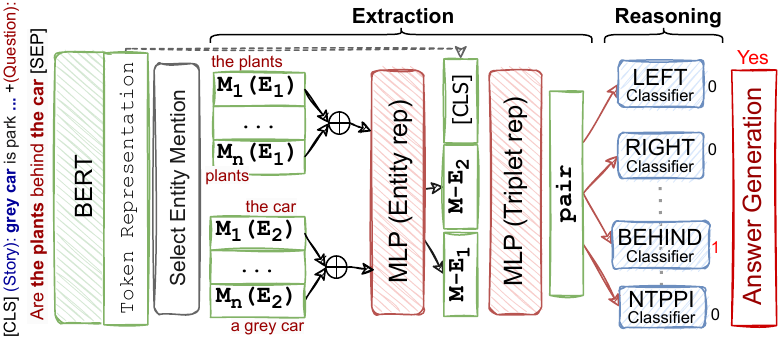}
	\caption{Model structure. 
First, entity mentions such as `plants' and `grey car' are selected from the \bert{} output and the entity representation is formed. Next, triplets like (`plants', `car', [CLS]) are generated and fed into the reasoning component. The collective output of all relation classifiers determines the relationships between each pair.
 *All hatched parts are trained end-to-end. The rest of the data is obtained from annotations or off-the-shelf modules.}
	\label{fig:sreqa-model}
    \end{subfigure}
    \begin{subfigure}{\linewidth}%
    \includegraphics[width=\linewidth]{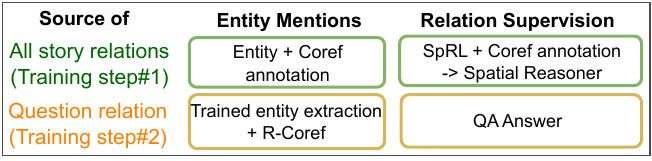}
	\caption{The source of supervision in each step of training. In step\#1, we train the model on all story relations, and in step\#2, we only train it on question relations.These modules and data are the same as the ones used in \pistaq{}.}
	\label{fig:sreqa-training}
    \end{subfigure}
    \caption{The \rtoe{} model with explicit neural layers to disentangle extraction and reasoning part.}
    \label{fig:sreqa}
\end{figure}

As another approach, we aim to explore a model that disentangles the extraction and reasoning parts inside a neural model.
Here, rather than directly predicting the answer from the output of PLMs~(as typically done in the QA task), we introduce explicit layers on top of PLM outputs. These layers are designed to generate representations for entities and pairs of entities, which are then passed to neural layers to identify all relations.
We call this model \textbf{\sreqa{}}\footnote{\textbf{S}patial \textbf{R}elation \textbf{E}xtraction for \textbf{QA}}, which is an end-to-end spatial relation extraction model designed for QA. Figure~\ref{fig:sreqa} illustrates the structure of this model.

In this model, we first select the entity mentions~($M_j(E_1)$) from the \bert{} tokens representation and pass it to the extraction part shown in Figure~\ref{fig:sreqa-model}.
Next, the model computes entity representation~($M(E_1)$) by summing the \bert{} token representations of all entity's mentions and passing it to an MLP layer. 
Then for each pair of entities,
a triplet is created by concatenating the pair's entities representations and the \bert{} $[CLS]$ token representation. This triplet is passed through an MLP layer to compute the final pair representations.
Next, in the reasoning part in Figure~\ref{fig:sreqa-model}, for each relation type in Table~\ref{tab:spatial_relations}, we use a binary 2-layer MLP classifier to predict the probability of each relation between the pairs.
We remove the inconsistent relations by selecting one with a higher probability at inference time, e.g., $\operatorname{LEFT}$ and $\operatorname{RIGHT}$ cannot be true at the same time. The final output is a list of all possible relations for each pair.
This model is trained using the summation of Focal loss~\cite{lin2017focal} of all relation classifiers. 


We train \sreqa{} in two separate steps.
In the first step, the model is trained on a relation extraction task which extracts \textit{all direct and indirect relations} between \textit{each pair of entities} only from stories. The top row of Figure~\ref{fig:sreqa-training} shows the annotation and modules employed in this step to gather the necessary supervision. 
We use the entity and coreference annotation to select the entity mentions from the \bert{} output. 
To compute the relations supervision for each pair of entities, we employ the spatial reasoner from \pistaq{} and apply it to the direct relations~(triplets) from the \sprl{} annotation, which are connected to each other by coreference annotations. This step of training is only feasible for datasets with available \sprl{} and coreference annotations. 

In the next step, we further train \sreqa{} on extracting \textit{questions relation} using QA supervision.
As shown in the bottom row of Figure~\ref{fig:sreqa-training}, we employ the trained spatial role extraction model used in \pistaq{} to identify the entities in the question and use R-Coref to find the mentions of these entities in the story. 
To obtain the relation supervision, we convert the question answers to relation labels. In FR, the label is similar to the actual answer, which is a list of all relations.
In YN, the question relation is converted to a label based on the Yes/No answer. For example, in Figure~\ref{fig:sreqa-model}, the question relation is `BEHIND,' and the answer is Yes, so the label for the BEHIND classifier is $1$.

We evaluate the \sreqa{} model's performance in predicting the accurate answers of the test set's questions same as training step 2.

%% file: 04-experiments.tex
\section{Experiments}

\subsection{Datasets}
\noindent\textbf{\spartqa{}} is an SQA dataset in which examples contain a story and multiple YN\footnote{We ignore ``Dont know'' answers in YN question and change them to No} and FR questions that require multi-hop spatial reasoning to be answered. The stories in this dataset describe relations between entities in a controlled~(toy task) environment.  
This dataset contains a large synthesized part, \auto{}, and a more complex small human-generated subset, \human{}. All stories and questions in this dataset also contain \sprl{} annotations.

\noindent\textbf{\spartun{}} 
is an extension of \spartqa{} with YN and FR questions containing \sprl{} annotations. Compared to \spartqa{}, the vocabulary and relation types in this dataset are extended, and it covers more relation types, rules, and spatial expressions to describe relations between entities. 

\noindent\textbf{\resq{}} is an SQA dataset with Yes/No questions over the human-generated text describing spatial relations in real-world settings.
The texts of this dataset are collected from \textbf{\msprl{}} dataset~\cite{kordjamshidi2017clef}~(See Figure~\ref{fig:resq}), which describe some spatial relations in pictures of ImageCLEF~\cite{grubinger2006iapr}. Also, the \msprl{} dataset already contains some \sprl{} annotations. To answer some of the questions in this dataset, extra spatial-commonsense information is needed~(e.g., a roof is on the \textit{top} of buildings).

\subsection{Model Configurations \& Baselines}

We compare the models described in section~\ref{sec:proposed_model} with the following baselines. 

\noindent \textbf{Majority Baseline}: This baseline selects the most frequent answer(s) in each dataset.

\noindent\textbf{GT-\etor{}:} 
This model uses ground truth~(GT) values of all involved modules in \etor{} to eliminate the effect of  error propagation in the pipeline. 
This baseline is used to evaluate the alignments between the questions and story entities and the reasoning module in solving the QA task.
It also gives an upper bound for the performance of the pipeline model, as the extraction part is perfect.



\noindent \textbf{\bert{}}: We select \bert{} as a candidate PLM that entangles the extraction and reasoning steps.
In this model, the input of the ``question$+$story'' is passed to the \bert{}, and the $[CLS]$ representation is used to do the answer classification. 


\noindent \textbf{GPT3.5}: GPT3.5~\cite{brown2020language} baselines~(GPT3.5 text-davinci-003) is selected as a candidate of generative larger language models which already passes many SOTAs in reasoning tasks~\cite{bang2023multitask, kojima2022large}. We use $Zero\_shot$ and $Few\_shot$~(In-context learning with few examples) settings to evaluate this model on the human-generated benchmarks. We also evaluate the Chain-of-Thoughts~(CoT) prompting method~\cite{wei2022chain} to extend the prompts with manually-written reasoning steps. The format of the input and some prompt examples are presented in Appendix~\ref{appendix:LLM}.


Transfer learning has already demonstrated significant enhancements in numerous deep learning tasks~\cite{so2022transfer,rajaby-faghihi-kordjamshidi-2021-time}. Thus, when applicable, we further train models on \spartun{} synthetic data shown by ``*''. 
The datasets' examples and statistics and more details of the experimental setups and configurations are provided in Appendix~\ref{appendix:statistics} and \ref{sec:setup}. All codes are publicly available at \href{https://github.com/RshNk73/PistaQ-SREQA}{https://github.com/RshNk73/PistaQ-SREQA}.

%% file: 05-results.tex
\section{Results and Discussion}



\begin{table}[t]
\resizebox{\columnwidth}{!}{%
\begin{tabular}{|l|l|l|}
\hline
\textbf{Model}                                     & \textbf{Supervisions}            & \textbf{Rule-based Modules} \\ \hline
BERT                                      & QA                      & \textbf{-}                     \\
GPT3.5$^{zero\_shot}$                       & -                       & \textbf{-}                     \\
GPT3.5$^{few\_shot}$                        & QA(8 ex)                & \textbf{-}                     \\
GPT3.5$^{few\_shot}+$CoT                    & QA(8 ex) $+$ CoT        & \textbf{-}                     \\ \hline
\berteq{}                & QA $+$SpRL(S)           & \textbf{-}                     \\ \hline
\sreqa{}                 & QA $+$SpRL(all)$+$Coref & Reasoner, R-Coref     \\
\sreqa{}*                & QA $+$ SpRL(Q)          & R-Coref            \\ \cline{1-2}
\pistaq{}                & SpRL(all) $+$ Coref     & Reasoner, R-Coref      \\
\pistaq{}$^{zero\_shot}$ & -                       & Reasoner, R-Coref     \\ \hline
\end{tabular}%
}
\caption{
The list of annotations from the target benchmarks and rule-based modules employed in each model. 
We use a quarter of \sprl{} annotations to train the modules on auto-generated benchmarks. 
S: Stories, Q: Questions, All: Stories$+$Questions.}
\label{tab:supervision}
\end{table}




Here, we discuss the influence of disentangling extraction and reasoning manifested in \pistaq{} and \sreqa{} models compared to various end-to-end models with no explicit design for these modules, such as \bert{}, \berteq{}, and GPT3.5. 
Table~\ref{tab:supervision} shows the list of these models with 
the sources of their supervision as well as extra off-the-shelf or rule-based modules employed in them. 








Since the performance of extraction modules, Spatial Role Labeling~(\sprl{}) and Coreference Resolution~(Coref), directly contribute to the final accuracy of the designed models, we have evaluated these modules and reported the results in Table~\ref{tab:sprl-result}. We choose the best modules on each dataset for experiments. For a detailed discussion on the performance of these modules, see Appendix~\ref{appendix:sprl-modules}.

\begin{table}[ht]
\scriptsize
\centering
\resizebox{\columnwidth}{!}{\begin{tabular}{|lcccc|}
\hline
\textbf{Dataset}& \textbf{Coref}&\textbf{SRole} & \textbf{SRel} & \textbf{SType} \\ \hline
\msprl{} & - & 88.59& 69.12& 19.79\\
\msprl{}* & - & 88.03 & 71.23 & 23.65\\ \hline
\multirow{2}{*}{\textsc{Human}} & \multirow{2}{*}{82.16} & \multirow{2}{*}{55.8}& S: 57.43& 43.79 \\
 &  & & Q: 52.55& 39.34 \\
\multirow{2}{*}{\textsc{Human}*} & \multirow{2}{*}{81.51}&\multirow{2}{*}{72.53} & S: 60.24& 48.74\\ 
 &  & & Q: 61.53&  48.07\\ \hline
\multirow{2}{*}{\spartqa{}} & \multirow{2}{*}{99.83} & \multirow{2}{*}{99.92} &S: 99.72& 99.05\\ 
 &  & & Q: 98.36& 98.62 \\ \hline
\multirow{2}{*}{\spartun{}} & \multirow{2}{*}{99.35} & \multirow{2}{*}{99.96}& S: 99.18& 98.57\\ 
 &  & & Q: 97.68& 98.11 \\ \hline

\end{tabular}}

\caption{Performance of the extraction modules. Q: question. S: stories. \textsc{HUMAN}: \human{}. \spartqa{}: \auto{}. *Further pretraining modules on \spartun{}.
We report macro F1 for \sprl{} and the accuracy of the Coref modules. }
\label{tab:sprl-result}

\end{table}




\subsection{Result on Controlled Environment} 

Table~\ref{tab:results_auto} shows the performance of models on two auto-generated benchmarks, \spartun{} and \auto{}.
We can observe that \pistaq{} outperforms all PLM baselines and \sreqa{}.
This outcome first highlights the effectiveness of the extraction and symbolic reasoning pipeline compared to PLMs in addressing deterministic reasoning within a controlled environment. Second, it shows that disentangling extraction and reasoning as a pipeline works better than explicit neural layers in SQA with a controlled environment. 
The complexity of these environments is more related to conducting several reasoning steps and demands accurate logical computations where a rule-based reasoner excels. Thus, the result of \pistaq{} with a rule-based reasoner module is also higher than \sreqa{} with a neural reasoner.

The superior performance of \pistaq{} over \bert{} suggests that \sprl{} annotations are more effective in the \pistaq{} pipeline than when utilized in \berteq{} in the form of QA supervision. 
Note that the extraction modules of \pistaq{} achieve perfect results on auto-generated benchmarks while trained only on a quarter of the \sprl{} annotations as shown in Table~\ref{tab:results_auto}.
However, \berteq{} uses all the original dataset questions and extra questions created from the full \sprl{} annotations.

\

\begin{table}[t]

    \centering
    \resizebox{\linewidth}{!}{
    \begin{tabular}{|c|l|c|c|c|c|}
        \hline
        
        \multirow{2}{*}{\#}&
        \multirow{2}{*}{Models}&\multicolumn{2}{c|}{\multirow{1}{*}{\spartun{}}} & \multicolumn{2}{c|}{\auto{}}
        \\ \cline{3-6}
        
        &&YN&FR&YN&FR
        \\ 
         \hline
          1&Majority baseline &53.62&14.23&51.82 & 44.35
          \\ 
          2&GT-\etor{}&99.07&99.43&99.51
            &98.99
            \\
          \hline
         3&BERT &91.80&91.80&84.88&94.17
         \\
        4&BERT-EQ&90.71&N/A&85.60&N/A\\
         
        \hline
          5&\sreqa{}&88.21&83.31& 85.11&86.88
\\
\hline
        6&\etor&\textbf{96.37}&\textbf{94.52}&\textbf{97.56}&\textbf{98.02}
          \\
\hline

    \end{tabular}
    }

        \caption{Results on auto-generated datasets. We use the accuracy metric for both YN and FR questions.}
    \label{tab:results_auto}
\end{table}


Table~\ref{tab:results_human} demonstrates the results of models on \human{} with a controlled environment setting. 
As can be seen, our proposed pipeline, \pistaq{}, outperforms the PLMs by a margin of 15\% on YN questions, even though the extraction modules, shown in Table~\ref{tab:sprl-result}, perform low. This low performance is due to the ambiguity of human language and smaller training data.
We also evaluate \pistaq{} on \human{} FR questions using Macro\_f1 score on all relation types. \pistaq{} outperforms all other baselines on FR questions, except for \bertspartun{}.

There are two main reasons behind the inconsistency in performance between YN and FR question types. The first reason is the complexity of the YN questions, which goes beyond the basics of spatial reasoning and is due to using quantifiers~(e.g., all circles, any object).
While previous studies have demonstrated that PLMs struggle with quantifiers~\cite{mirzaee-etal-2021-spartqa}, the reasoning module in \pistaq{} can adeptly handle them without any performance loss.
Second, further analysis indicates that \pistaq{} predicts `No' when a relationship is not extracted, which can be correct when the answer is `No'. However, in FR, a missed extraction causes a false negative which decreases F1 score.

\begin{table}[t]

    \centering
    \resizebox{\linewidth}{!}{
    \begin{tabular}{|c|l|c|c|c|c|}
        \hline
        
        
        \multirow{2}{*}{\#}&\multirow{2}{*}{Models}&YN&\multicolumn{3}{c|}{FR}\\ \cline{3-6}
        &&Acc&P&R&F1
        \\ 
         \hline
          1&Majority baseline &52.44&29.87&14.28&6.57
          \\ 
          2&GT-\etor{}&79.72&96.38&66.04&75.16
            \\
          \hline
         3&BERT  &51.74&30.74&30.13&28.17
         \\
         4&\bert{}* &48.95&60.96&49.10& \textbf{50.56}
         \\ 
         5&GPT3.5$^{Zero\_shot}$ &45.45&40.13&22.42&16.51
         \\ 
        6&GPT3.5$^{Few\_shot}$ &60.13&45.20&54.10& 44.28
         \\ 
         7&GPT3.5$^{Few\_shot}+$CoT &62.93&57.18&37.92&38.47
         \\ 
        \hline
        8&BERT-EQ &50.34&-&-&- \\
        9&BERT-EQ* &45.45&-&-& -\\
        \hline


        10&\sreqa{}&53.23&15.68&13.85&13.70
\\
          11&\sreqa{}*&46.96&18.70&25.79& 24.61
\\
\hline
        12&\etor&\textbf{75.52}&72.11&35.93&46.80
          \\
\hline

    \end{tabular}
    }
        \caption{Results on \human{}. We use accuracy on YN questions and average Precision~(P), Recall~(R), and Macro-F1 on FR question types. *Using \spartun{} supervision for further training.}
    \label{tab:results_human}
\end{table}

\begin{table}[t]
\tiny
\centering

\resizebox{\columnwidth}{!}{%
\begin{tabular}{|c|l|c|}
\hline
\# & Models  & Accuracy \\ \hline 
1 & Majority baseline & 50.21 \\ \hline 
2 & BERT & 57.37 \\
3 & BERT*$^{Zero\_shot}$ &  49.18 \\
4 & \bertspartun{} &  63.60 \\
5 & GPT3.5$^{Zero\_shot}$ & 60.32 \\
6 & GPT3.5$^{Few\_shot}$ &  65.90 \\ 
7 & GPT3.5$^{Few\_shot}+$CoT & 67.05 \\ 
\hline
8 & BERT-EQ & 56.55 \\
9 & BERT-EQ*$^{Zero\_shot}$ & 51.96 \\
10 & BERT-EQ* & 61.47 \\ \hline 
11 & \sreqa{} &  53.15 \\
12 & \sreqa{}*$^{Zero\_shot}$ &  53.32 \\
13 & \sreqa{}* &  \textbf{69.50} \\ \hline 
14 & \etor$^{\msprl{}}$  & 41.96 \\
15 & \etor~$^{\spartun{}+\msprl{}}$  & 47.21 \\ \hline
16 & Human  & 90.38 \\ \hline
\end{tabular}%
}
\caption{Result on \resq{}. *Further training on \spartun{}. The $Zero\_shot$ refers to evaluation without further training on \resq{} or \msprl{} training data.}
\label{tab:results_resq}
\end{table}


\begin{figure*}[t]
    \centering	\includegraphics[width=\linewidth]{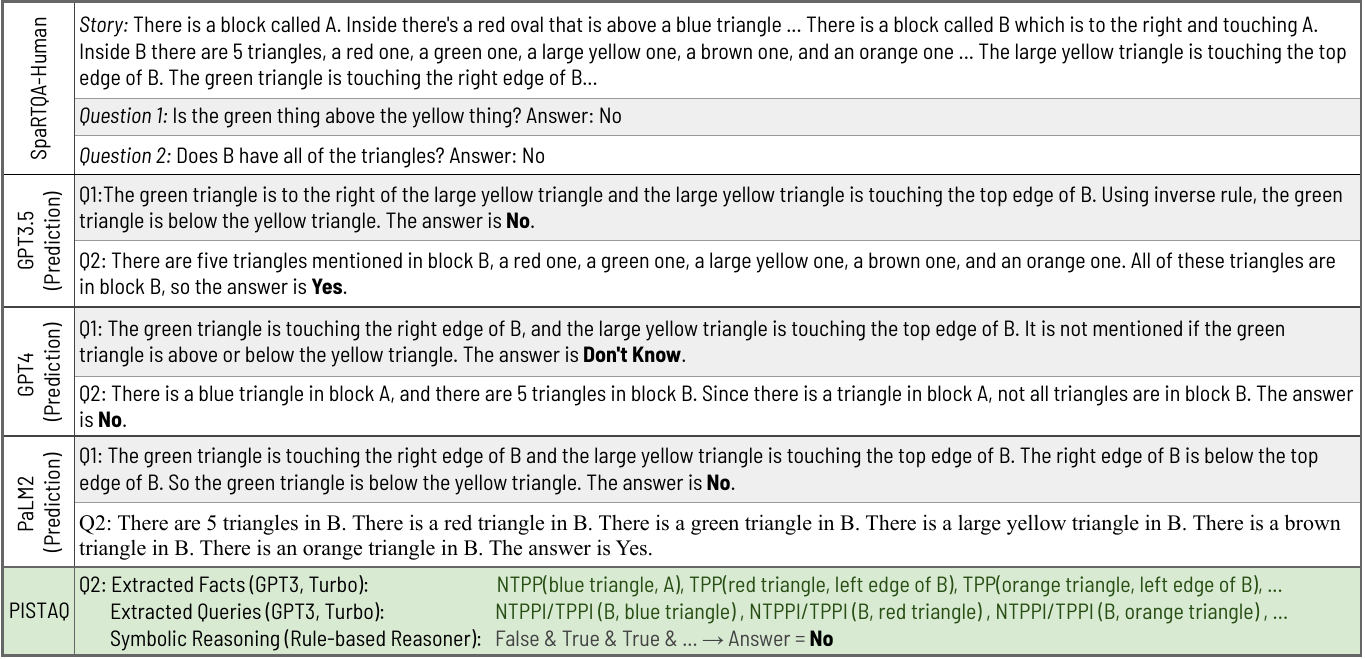}
		\caption{An example of Large Language Models~(LLMs) prediction on SQA task evaluated with CoT prompting. The last row shows an example of using GPT3.5-Turbo for information extraction in \pistaq{}. See Appendix~\ref{appendix:LLM} for $zero\_shot$ examples.}
		\label{fig:LLM}
\end{figure*}

\subsection{Results on Real-world Setting}
We select \resq{} as an SQA dataset with realistic settings and present the result of models on this dataset in Table~\ref{tab:results_resq}. 

To evaluate \pistaq{} on \resq{}, we begin by adapting its extraction modules through training on the corresponding dataset. We train the \sprl{} modules on both \msprl{} and \spartun{}, and the performance of these models is presented in Table~\ref{tab:sprl-result}.
As the \msprl{} dataset lacks coreference annotations, we employ the model trained on \spartun{} for this purpose. 
Rows 14 and 15 in Table~\ref{tab:results_resq} show the performance of the \pistaq{} on \resq{} is inferior compared to other baselines.
To find the reason, we analyze the first 25 questions from the \resq{} test set. 
We find that 18 out of 25 questions required spatial commonsense information and  cannot be answered solely based on the given relations in the stories. From the remaining 7 questions, only 2 can be answered using the \sprl{} annotations provided in the \msprl{} dataset. Some examples of this analysis are provided in Appendix~\ref{appendix:msprl-missed}.
Hence, the low performance of \pistaq{} is attributed to first the absence of integrating commonsense information in this model
and, second, the errors in the extraction modules, which are propagated to the reasoning modules.

As shown in Table~\ref{tab:results_resq}, the best result on \resq{} is achieved by \sreqa{}* model. Compared to \sreqa{}, \sreqa{}* is trained on \spartun{} instead of \msprl{}\footnote{As mentioned, we use the \msprl{} annotation for \resq{} dataset.} in the first step of the training.
\msprl{} lacks some \sprl{} and coreference annotations to answer \resq{} questions.
In the absence of this information, 
collecting the supervision for the first phase of training
results in a significant number of missed relations.
Therefore, as shown in row 11 of Table~\ref{tab:results_resq}, employing \msprl{} in the first training phase decreases the performance while replacing it with \spartun{} in \sreqa{}* significantly enhances the results.



\sreqa{}* surpasses the PLMs trained on QA and QA+\sprl{} annotation, showcasing the advantage of the design of this model in utilizing QA and \sprl{} data within explicit extraction layers and the data preprocessing.
Also, the better performance of this model compared to \pistaq{} demonstrates how the end-to-end structure of \sreqa{} can handle the errors from the extraction part and also can capture some rules and commonsense knowledge from \resq{} training data that are not explicitly supported in the symbolic reasoner.

In \textit{conclusion}, compared to PLMs, disentangling extraction and reasoning as a pipeline indicates superior performance in deterministic spatial reasoning within controlled settings.
Moreover, explicitly training the extraction module proves advantageous in leveraging \sprl{} annotation more effectively compared to using this annotation in QA format in the end-to-end training.
Comparison between disentangling extraction and reasoning as a pipeline and incorporating them within an end-to-end model demonstrates that the end-to-end model performs better in realistic domains even better than PLMs.
The end-to-end architecture of this model effectively enhances the generalization in the real-world setting and addresses some of the limitations of rule coverage and commonsense knowledge.

\subsection{LLMs on Spatial Reasoning}

Recent research shows the high performance of LLMs with $zero$/$few\_shot$ setting on many tasks~\cite{chowdhery2022palm, brown2020language}.
However, \cite{bang2023multitask} shows that ChatGPT~(GPT3.5-Turbo) with $zero\_shot$ evaluation cannot perform well on SQA task using \human{} test cases.
Similarly, our experiments, as shown in
Tables~\ref{tab:results_human} and  \ref{tab:results_resq}, show the lower performance of GPT3.5~(davinci) with $zero$/$few\_shot$ settings compared to human and our models \pistaq{} and \sreqa{}.
Figure~\ref{fig:LLM}, shows an example of three LLMs, GPT3.5, GPT4 and PaLM2 on \human{} example\footnote{Due to the limited resources, we only use GPT4 and PaLM2 on a few examples to evaluate their performance on SQA tasks.}~(complete figure including $zero\_shot$ examples is presented in Appendix~\ref{appendix:LLM}). Although \citeauthor{wei2022chain} shows that using CoT prompting improves the performance of PaLM on multi-step reasoning task, its spatial reasoning capabilities still does not meet the expectation.  


\subsubsection{LLMs as Extraction Module in \pistaq{}}
\label{sec:llm-extraction}
\begin{figure}[t]
    \centering
    \includegraphics[width=\columnwidth]{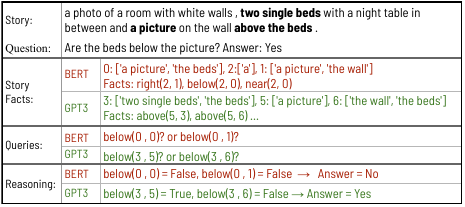}
    \caption{An example of using \bert{}-based \sprl{} and GPT3.5 as information extraction in \pistaq{} on a \resq{} example.}
    \label{fig:llm-pistaq}
\end{figure}


A recent study~\cite{shen2023large} shows that LLMs have a promising performance in information retrieval. 
Building upon this, we employ LLM, GPT3.5-Turbo with $few\_shot$ prompting to extract information from a set of \human{} and \resq{} examples that do not necessitate commonsense reasoning for answering. The extracted information is subsequently utilized within the framework of \pistaq{}.

The results, illustrated in the last row of Figure~\ref{fig:LLM}, highlight how the combination of LLM extraction and symbolic reasoning enables answering questions that LLMs struggle to address. Furthermore, Figure~\ref{fig:llm-pistaq} provides a comparison between the trained \bert{}-based \sprl{} extraction modules and GPT3.5 with $few\_shot$ prompting in \pistaq{}. It is evident that GPT3.5 extracts more accurate information, leading to correct answers. 
As we mentioned before, out of 25 sampled questions from \resq{}, only 7 can be solved without relying on spatial commonsense information.
Our experimental result shows that \pistaq{} using LLM as extraction modules can solve all of these 7 questions.


Based on these findings, leveraging LLMs in \pistaq{} to mitigate errors stemming from the \sprl{} extraction modules rather than relying solely on LLMs for reasoning can be an interesting future research direction. 
This insight emphasizes the importance of considering new approaches for incorporating explicit reasoning modules whenever possible instead of counting solely on black-box neural models for reasoning tasks.

%% file: 10-conclusion.tex
\section{Conclusion and Future Works}


We investigate the benefits of  disentangling the processes of extracting spatial information and reasoning over them. To this end, we devised a series of experiments utilizing PLMs for spatial information extraction coupled with a symbolic reasoner for inferring indirect relations.
The outcomes of our experiments provide noteworthy insights: 
(1) Our observations in controlled experimental conditions demonstrate that disentangling extraction and symbolic reasoning, compared to PLMs, enhances the models' reasoning capabilities, even with comparable or reduced supervision.
(2) Despite the acknowledged fragility of symbolic reasoning in real-world domains, our experiments highlight that employing explicit extraction layers and utilizing the same symbolic reasoner in data preprocessing enhances the reasoning capabilities of models.
(3) Despite the limitations of LLMs in spatial reasoning, harnessing their potential for information extraction within a disentangled structure of Extraction and Reasoning can yield significant benefits.
All of these results emphasize the advantage of disentangling the extraction and reasoning in spatial language understanding.





In future research, an intriguing direction is incorporating spatial commonsense knowledge using LLMs as an extraction module in the pipeline of extraction and reasoning.
Additionally, the model's applicability extends beyond spatial reasoning, making it suitable for various reasoning tasks involving logical rules, such as temporal or arithmetic reasoning.

%% file: 99-appendix.tex


\section{Statistic Information}
\label{appendix:statistics}

This section presents statistical information regarding dataset sizes and additional analyses conducted on the evaluation sets of human-generated datasets.

Table~\ref{tab:statistical} provides the number of questions in the training and evaluation sets of the SQA benchmarks.
Tables~\ref{tab:sprl_statistical} and \ref{tab:triplet_statistical} present the sentence and number of relation triplets within the \sprl{} annotation for each dataset, respectively.
Table~\ref{tab:msprl-statistic} illustrates a comprehensive breakdown of the size of Role and Relation sets in the \msprl{} dataset.

\begin{table}[th]
\small

\centering
\resizebox{\columnwidth}{!}{\begin{tabular}{|l|ccc|}
\hline
\textbf{Dataset} & \multicolumn{1}{l}{\textbf{Train}} & \multicolumn{1}{l}{\textbf{Dev}} & \multicolumn{1}{l|}{\textbf{Test}} \\ \hline
\auto{}~(YN) & 26152 &3860  & 3896   \\
\auto{}~(FR) & 25744 &3780  & 3797   \\ 
\multicolumn{1}{|l|}{\human{}~(YN)}& 162 & 51 & 143 \\
\multicolumn{1}{|l|}{\human{}~(FR)}& 149 & 28 & 77  \\ 
\cline{1-1}
\spartun{}~(YN) & 20334 &3152& 3193\\
\spartun{}~(FR) & 18400 &2818& 2830\\\cline{1-1}
\resq{}(YN) & 1008 & 333 & 610 \\
\hline
\end{tabular}}
\caption{Number of questions in training and evaluation sets of SQA benchmarks.}
\label{tab:statistical}
\end{table}

\begin{table}[th]

\centering
\resizebox{\columnwidth}{!}{\begin{tabular}{|l|ccc|}
\hline
\textbf{Dataset} & \multicolumn{1}{l}{\textbf{Train}} & \multicolumn{1}{l}{\textbf{Dev}} & \multicolumn{1}{l|}{\textbf{Test}} \\ \hline
\auto{}~(story) & 36420 &16214  & 16336   \\
\auto{}~(question) & 53488 &15092  & 15216   \\
\cline{1-1}
\human{}~(story)& 389 &213  & 584  \\
\human{}~(question) & 623 &190  & 549   \\
\cline{1-1}
\spartun{}~(story) & 68048 &9720& 10013\\
\spartun{}~(question) & 41177 &6355& 6340\\
\cline{1-1}
\msprl{}&  600 & -  & 613 \\ 
\hline
\end{tabular}}
\caption{Number of sentences in \sprl{} annotations of each benchmarks. To train models on the auto-generated benchmarks, we only use the quarter of training examples from \spartun{} and \auto{}.}
\label{tab:sprl_statistical}
\end{table}

\begin{table}[th]

\centering
\resizebox{\columnwidth}{!}{\begin{tabular}{|l|ccc|}
\hline
\textbf{Dataset} & \multicolumn{1}{l}{\textbf{Train}} & \multicolumn{1}{l}{\textbf{Dev}} & \multicolumn{1}{l|}{\textbf{Test}} \\ \hline
\auto{}~(story) & 159712 &22029  & 21957   \\
\auto{}~(question) & 232187 &34903  & 35011   \\
\cline{1-1}
\human{}~(story)& 176 &99  & 272  \\
\human{}~(question) & 155 &127  & 367   \\
\cline{1-1}
\spartun{}~(story) & 48368 &7031& 7191\\
\spartun{}~(question) & 38734 &5970& 6023\\
\cline{1-1}
\msprl{}& 761 & -  & 939  \\ 
\hline
\end{tabular}}
\caption{Number of triplets in \sprl{} annotations of each benchmarks.}
\label{tab:triplet_statistical}
\end{table}

\begin{table}[th]

    \centering
    \begin{tabular}{|l|l|l|l|}

        \hline & \text{ Train } & \text{ Test } & \text{ All } \\
        \hline \text{ Sentences } & 600 & 613 & 1213 \\
        \hline \text{ Trajectors } & 716 & 874 & 1590 \\
        \text{ Landmarks } & 612 & 573 & 1185 \\
        \text{ Spatial Indicators } & 666 & 795 & 1461 \\
        \text{ Spatial Triplets } & 761 & 939 & 1700 \\
        \hline

    \end{tabular}
    \caption{\msprl{} size~\cite{kordjamshidi2017clef}.}
    \label{tab:msprl-statistic}
\end{table}

\begin{figure}
    \centering
    \includegraphics[width=\linewidth]{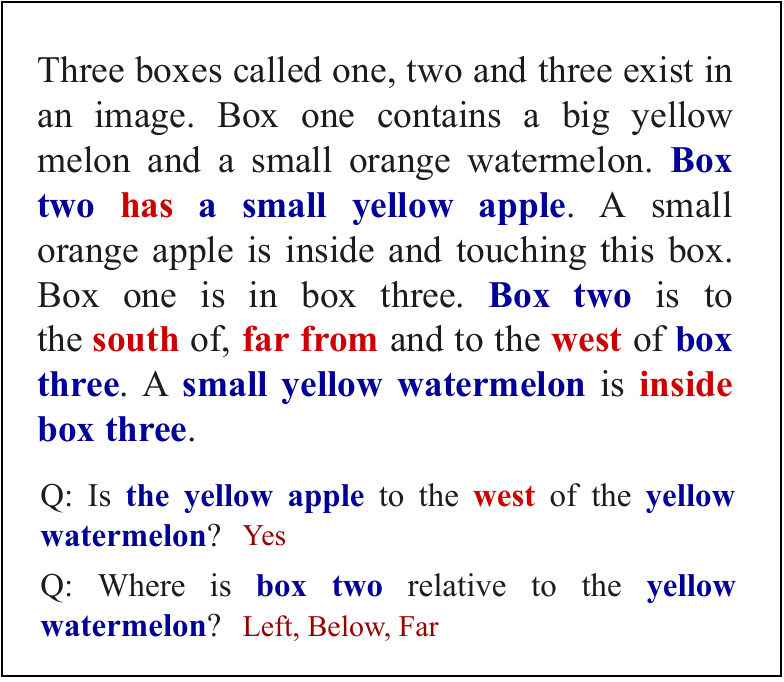}
    \caption{An example of \spartun{} dataset from \cite{spartun}.}
    \label{fig:spartun-ex}
\end{figure}


\subsection{Analyzing \human{} YN}
\label{appendix:yn-human}

We conducted additional evaluations on the superior performance of \pistaq{} over other baseline models on \human{} YN questions. As explained before, \pistaq{} tends to predict \textit{No} when information is not available, resulting in more \textit{No} and fewer \textit{Yes} predictions compared to other models, as presented in Table~\ref{tab:human-YN}. The number of true positive predictions for \pistaq{} is more than two other baselines, and as a result, it achieves higher accuracy.


\begin{table}[h]

\centering
\resizebox{0.9\columnwidth}{!}{%
\begin{tabular}{|l|l|l|l|}
\hline
Predictions/ Answer & Yes & No & No prediction \\ \hline
Ground Truth & 74 & 69 & - \\ \hline
BERT & 131 & 12 & - \\ \hline
BERT* & 89 & 54 & - \\ \hline
\pistaq{} & 43 & 97 & 3 \\ \hline
\end{tabular}%
}
\caption{Detailed information about the prediction of \pistaq{} and \bert{} on \human{} YN questions. ``No prediction'' is related to the \pistaq{} model when no correct \sprl{} extraction was made for the text of the question, and as a result, we have no answer prediction.}
\label{tab:human-YN}
\end{table}


\subsection{\sprl{} Annotations in \msprl{}}
\label{appendix:msprl-missed}


\begin{figure}[h]
    \centering
    \includegraphics[width=\linewidth]{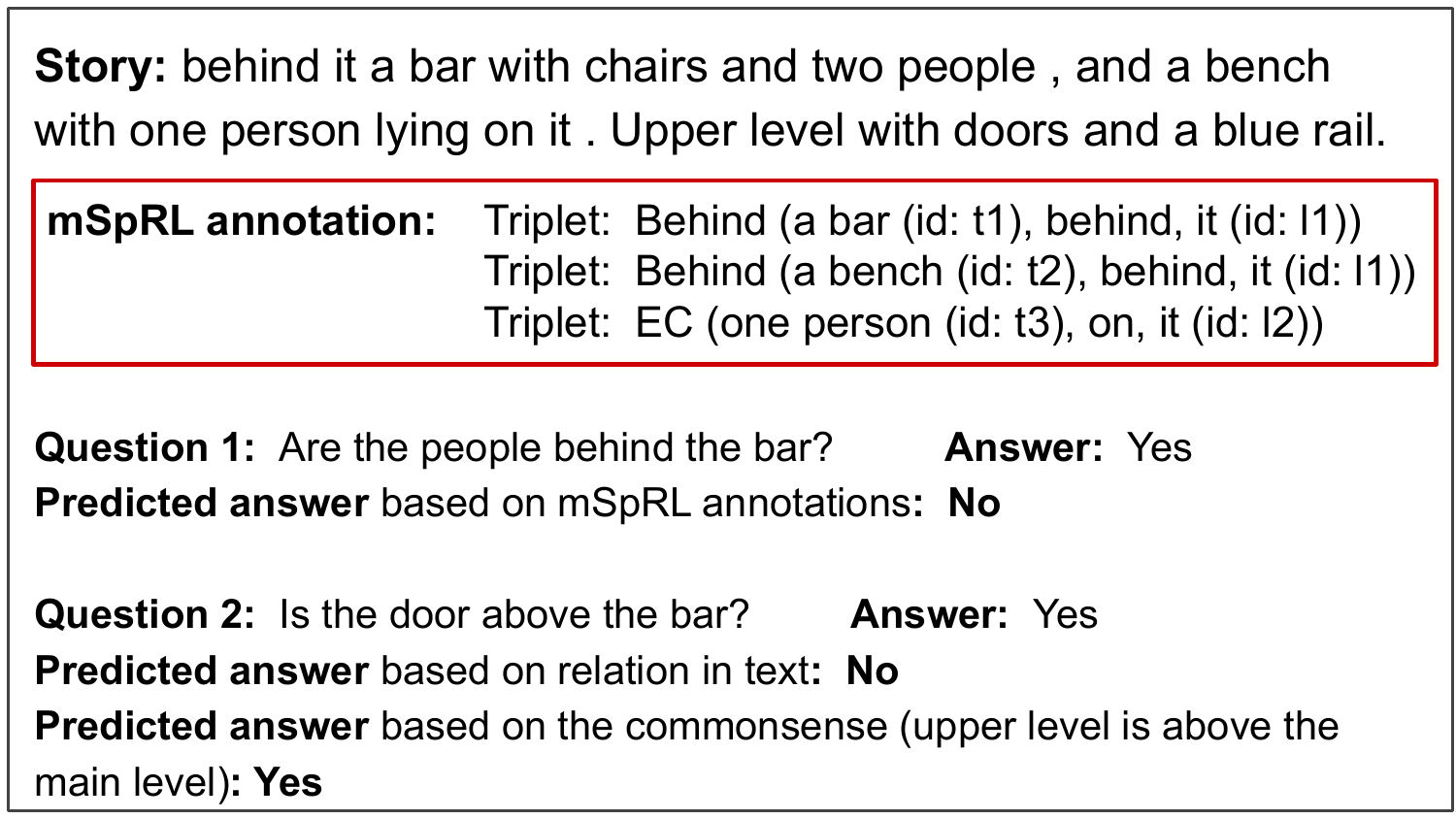}
    \caption{An example of the limitation of \msprl{} and coreference  annotation  to answer \resq{} question.
    The answer to the questions was predicted wrongly due to two main reasons. First, the missed commonsense knowledge in question 2, and second, the limited coverage of ground truth annotation in \msprl{} in question 2.}
    \label{fig:msprl-resq}
\end{figure}

\resq{} is built on the human-written context of \msprl{} dataset which includes \sprl{} annotations. However, using this annotation in \berteq{} and \rtoe{} models causes lower performance~(Check the result on Tablel~\ref{tab:sre-result}).  
Our analysis  shows that the \sprl{} annotations of \msprl{} are not fully practical in our work due to two main reasons:

\begin{enumerate}
    \item \textbf{Missed annotations:} As shown in Figure~\ref{fig:msprl-resq}, there are many missed annotations for each text, e.g., NTPP(bar, with, chair). 
    
    \item \textbf{No coreference :} The coreference annotation is not supported in this dataset, e.g., ``L2: it'' and ``T2: a bench'' are the same entity with different mentions, but they are mentioned with different ids. These missed coreferences result in fewer connections between entities and fewer inferred relations. 
\end{enumerate}



\section{Models and modules configuration}
\label{sec:setup}
We use the huggingFace\footnote{\url{https://huggingface.co/transformers/v2.9.1/model_doc/bert.html}} implementation of pretrained \bert{} base models, which have 768 hidden dimensions.  All models are trained on the training set, evaluated on the dev set, and reported the result on the test set. For training, we train the model until no changes happen on the dev set and then store and use the best model on the dev set. We use AdamW (\cite{loshchilov2017decoupled}), and learning rates from {$2\times 10^{-6}$, $2\times 10^{-5}$} (depends on the task and datasets) on all models and modules. 
For the extraction modules and the baselines, we used the same configuration and setting as previous works~\cite{spartun}.
For \rtoe{} models we use learning rates of {$2\times 10^{-5}$, $4\times 10^{-6}$} for \rtoe{}(story) and \rtoe{}(question) respectively. 
To run the models we use machine with Intel
Core i9-9820X (10 cores, 3.30 GHz) CPU and Titan RTX with NVLink as GPU.

For GPT3.5, we use Instruct-GPT, \textit{davinci-003}\footnote{from \url{https://beta.openai.com}}. The cost for running GPT3.5 on the human-generated benchmarks was 0.002\$ per 1k tokens. 
For GPT3.5 as information extraction, we use \textit{GPT3.5 turbo}~(a.k.a ChatGPT) with a cost of 0.0001\$ per 1k tokens.
We also use the GPT4 playground in OpenAI and PaLM2 playground to find the prediction of examples in Figure~\ref{fig:LLM-all}.

\input{99-module}

\section{\sreqa{} on All Story Relations}
\label{appendix:sreqa-story}

\begin{table}[h]
\tiny
\centering

\resizebox{\columnwidth}{!}{%
\begin{tabular}{|l|c|}
\hline
Datasets & F1 on \sreqa{} \\ \hline
\spartun{} & 96.37 \\ \hline
\auto{} & 97.78 \\ \hline
\human{} & 23.79 \\ \hline
\msprl{}~(Used in \resq{}) & 16.59 \\ \hline
\end{tabular}%
}
\caption{The result of \sreqa{} model only trained on all story relations of the SQA datasets.}
\label{tab:sre-result}
\end{table}


Table~\ref{tab:sre-result} displays the results of the \sreqa{} model trained and tested solely on all the story's relation extraction parts~(step 1). During the evaluation, we also possess the same data preprocessing and gather annotations of all relations between stories' entities and select the best model based on performance on the development set.

Notably, the performance on the human-generated datasets, \human{} and \resq{}, is significantly lower compared to the auto-generated datasets. As discussed in \label{appendix:msprl-missed}, the \msprl{} datasets contain missed annotations, resulting in the omission of several relations from the stories' entities and incomplete training data for this phase. Similarly, the \human{} \sprl{} annotation, as discussed in Appendix~\ref{appendix:sprl-modules}, exhibits some noise, particularly in coreference annotation, leading to similar issues as observed in \msprl{} regarding annotation of all story relations.

Consequently, this reduced performance in all story relation extraction impacts the overall performance of the main \sreqa{} model trained using two steps; however, as illustrated in the results of \sreqa{}* in Table\ref{tab:results_resq}, which utilizes Spartun instead of \msprl{} for training on all story's relations, the performance substantially improves on the \resq{} dataset.

\section{Large Language Models~(LLMs)}
\label{appendix:LLM}
\begin{figure*}[b]
    \centering
    \includegraphics[width=\linewidth]{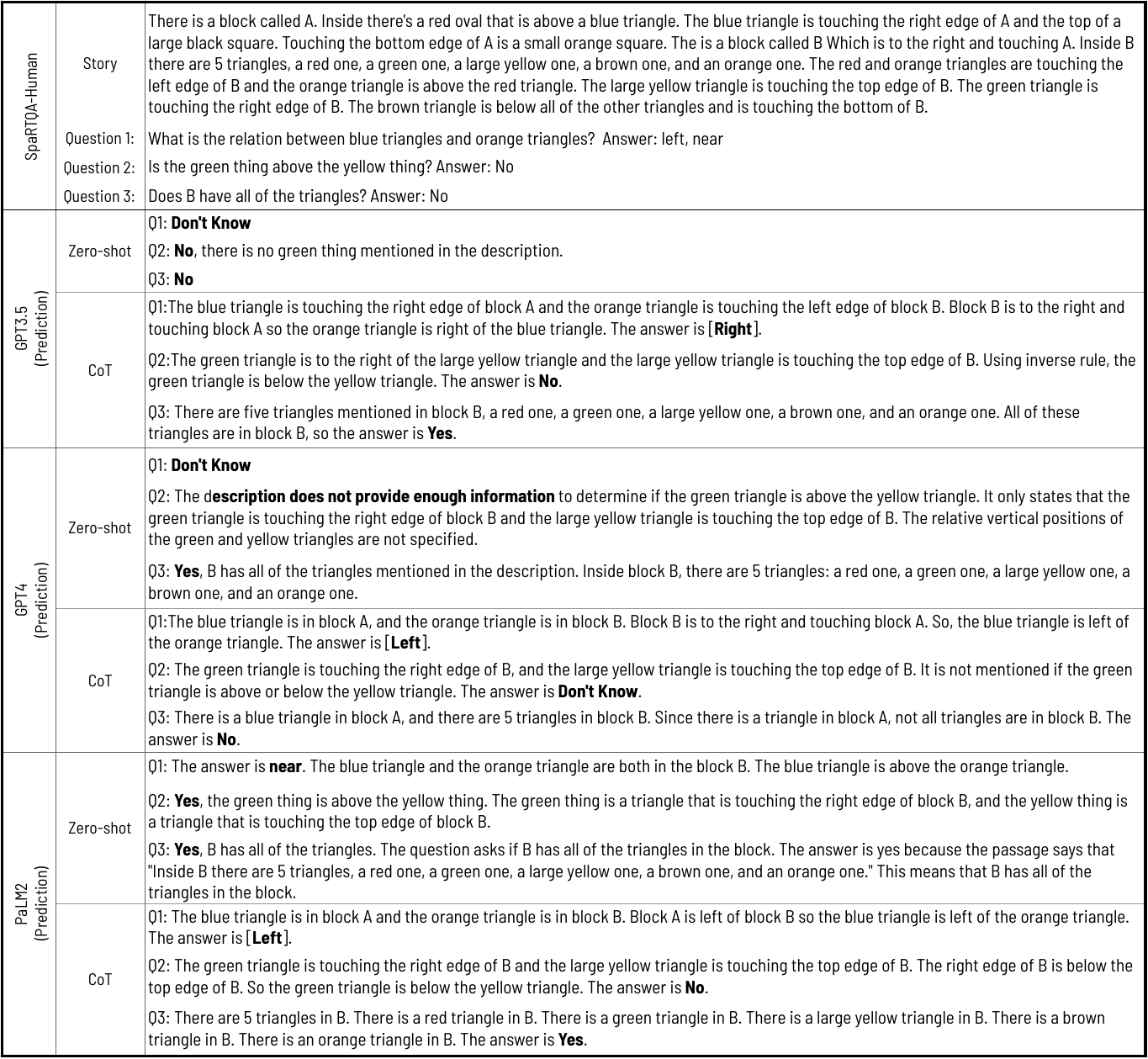}
    \caption{An example of three Large Language Models~(LLMs), GPT3.5~(davinci-text-003), GPT4, and PaLM2  prediction on SQA task evaluated with $zero\_shot$ and CoT prompting.}
    \label{fig:LLM-all}
\end{figure*}

Figure~\ref{fig:LLM-all} presents examples showcasing predictions made by three Large Language Models (LLMs): GPT3.5-DaVinci, GPT4, and PaLM2, on a story from the \human{} dataset. These examples demonstrate that while these models, specifically GPT4 and PaLM2, excel in multi-hop reasoning tasks, solving spatial question answering remains a challenging endeavor.

To evaluate the LLMs' performance on spatial reasoning, we use $Zero\_shot$, $Few\_shot$, and $Few\_shot+$CoT. 
In the $Zero\_shot$ setting, the prompt given as input to the model is formatted as ``Context: story. Question: question?'' and the model returns the answer to the question. 

\begin{figure*}[t]
    \centering
    \includegraphics[width=\linewidth]{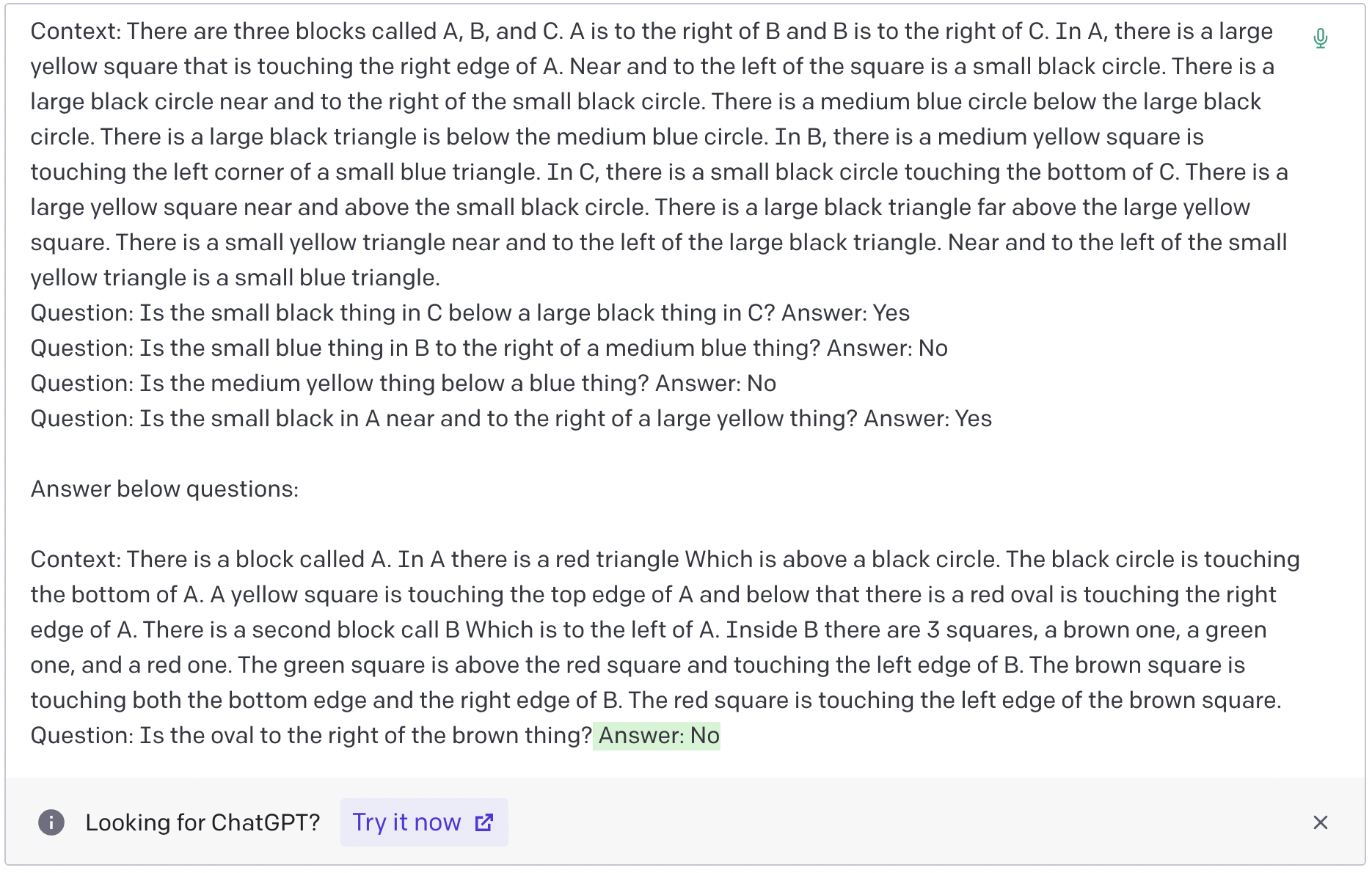}
    \caption{Example of the input for GPT3.5($Few\_shot$). The $Zero\_shot$ setting is the same just do not have the first training example.}
    \label{fig:gpt3}
\end{figure*}

In the $Few\_shot$ setting, we add two random examples from the training data with a story, all its questions, and their answers. Figure~\ref{fig:gpt3} depicts a prompt example for \human{} YN questions, passed to GPT3.5.

\begin{figure*}[h]
    \centering
    \includegraphics[width=\linewidth]{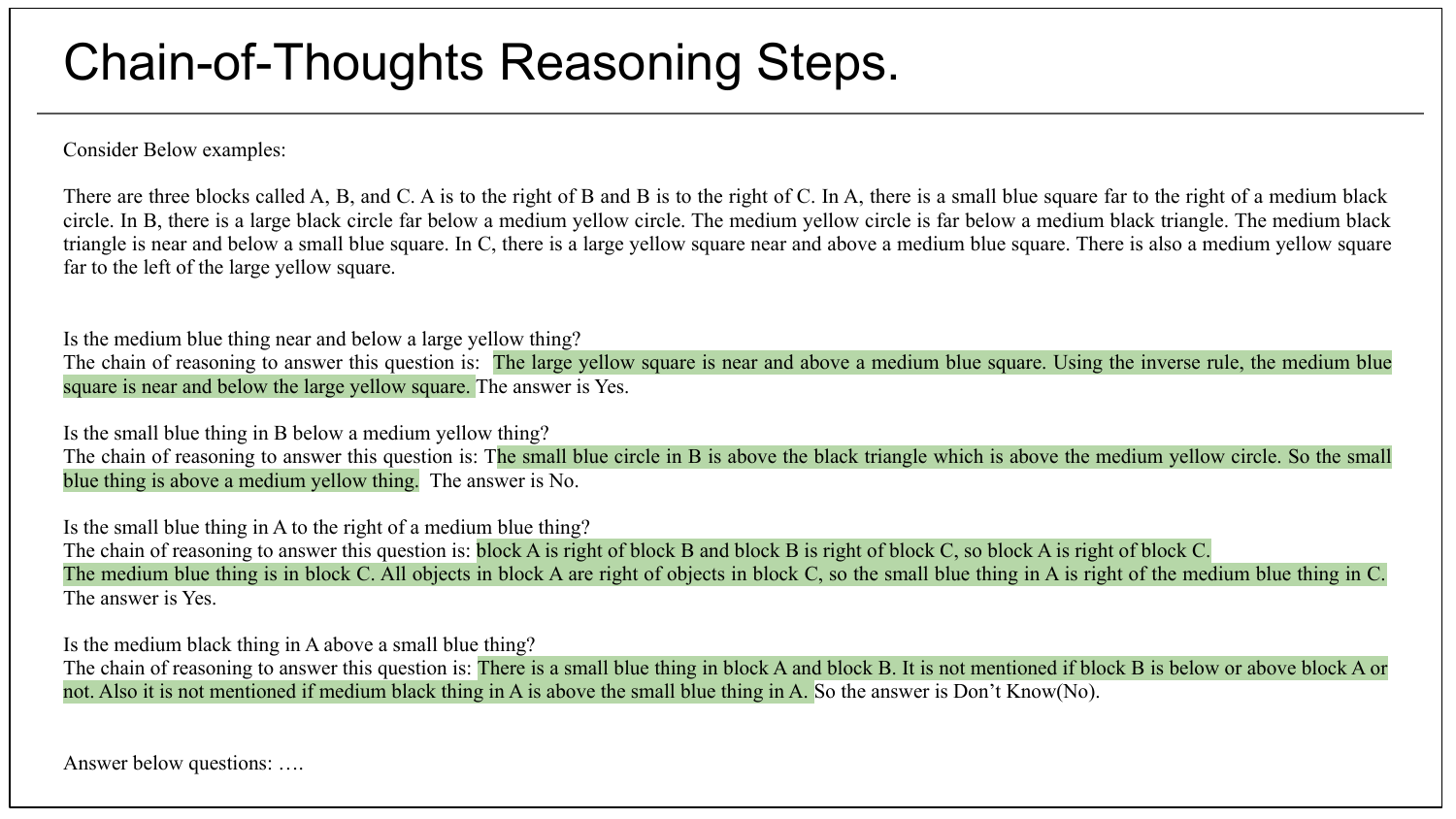}
    \caption{Example of the input for GPT3.5($Few\_shot+$Cot) with human-written Chain-of-Thoughts.}
    \label{fig:cot}
\end{figure*}

For $Few\_shot+$CoT, we use the same idea as ~\cite{wei2022chain} and manually write the reasoning steps for eight questions (from two random stories). The input then is formatted as ``Context: story. Question: CoT. Answer. Asked Context: story. Question:  question?''. Figure~\ref{fig:cot} shows an example of these reasoning steps on \resq{}.

\subsection{LLMs for Information Extraction}


As discussed in Section~\ref{sec:llm-extraction}, we utilize LLM, GPT3.5-Turbo, for information extraction from human-generated texts. The extraction process encompasses Entity, Relation, Relation Type, and coreference extractions from the story, as well as entity and relation extraction from the question. Additionally, LLM is employed to identify mentions of question entities within the text.

\begin{figure*}[ht]
    \centering
    \includegraphics[width=\linewidth]{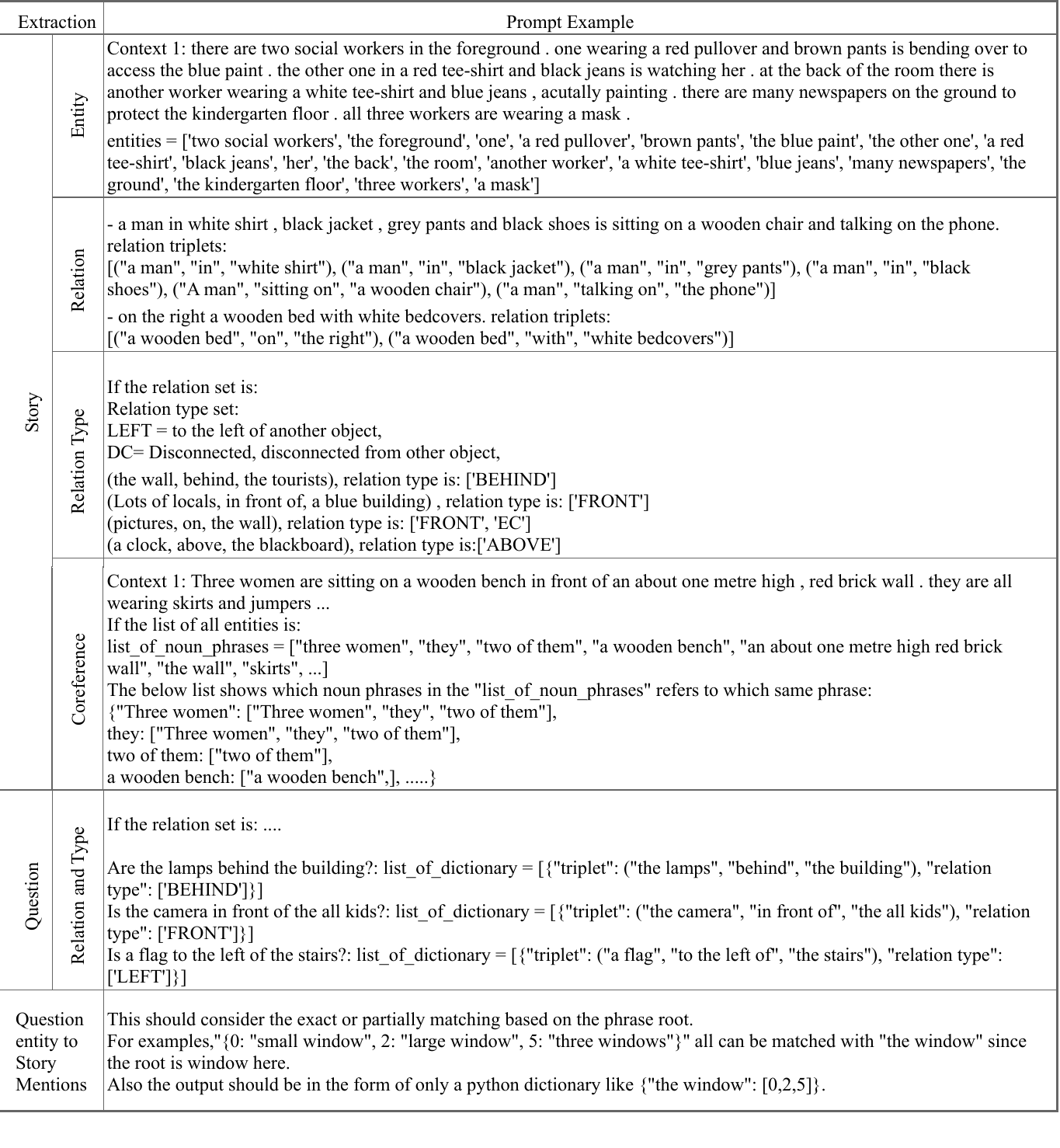}
    \caption{The example of prompts used for LLMs~(GPT3.5-Turbo) in information extraction.}
    \label{fig:llm-extraction}
\end{figure*}

We construct multiple manually crafted prompt examples for each extraction task, as depicted in Figure~\ref{fig:llm-extraction}. Subsequently, the extracted information is inputted into the reasoner module of \pistaq{} to compute the answers.

\begin{figure}
    \centering
    \includegraphics[width=\linewidth]{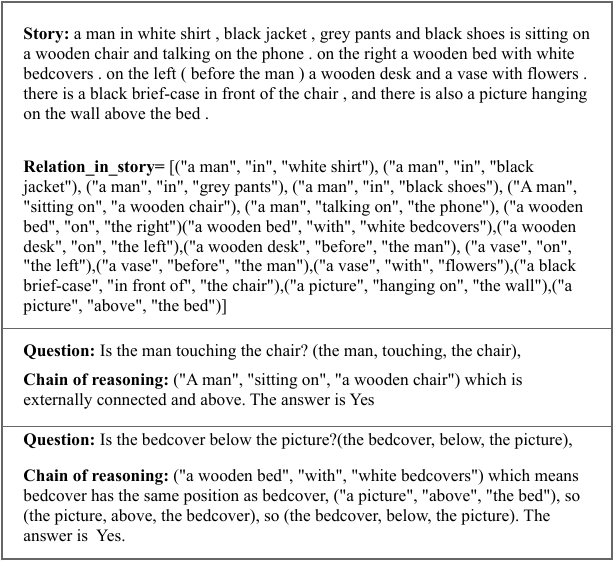}
    \caption{We employ LLMs in both extraction and reasoning tasks, but in a disentangled manner. Initially, we extract information using LLMs, and subsequently incorporate this extracted information into the prompt alongside written CoTs based on the extracted data. }
    \label{fig:LLM-reasoning}
\end{figure}

In addition to our experiment, we attempted to incorporate LLMs as neural spatial reasoners but in a pipeline structure of extraction and reasoning.
To do so, as illustrated in Figure~\ref{fig:LLM-reasoning}, we add the extracted information of LLM with the written CoTs based on this extracted information to the prompt of a GPT3.5-DaVinci. 
The results, however, become even lower (62.62\%) compared to GPT3.5-CoT with the main text (67.05\%) when evaluated on the \resq{} dataset. This outcome highlights the superior ability of LLMs to capture information from natural language compared to structured data without fine-tuning.


%% file: 99-module.tex
\section{Extraction and Reasoning Modules}
\label{appendix:sprl-modules}

Here, we discuss each module used in \etor{} and their performance including the \textit{Spatial Role Labeling} (\sprl{}), \textit{Coreference Resolution}, and  \textit{Spatial reasoner}. 


 \begin{figure}[ht]
    \centering
    \includegraphics[width=\linewidth]{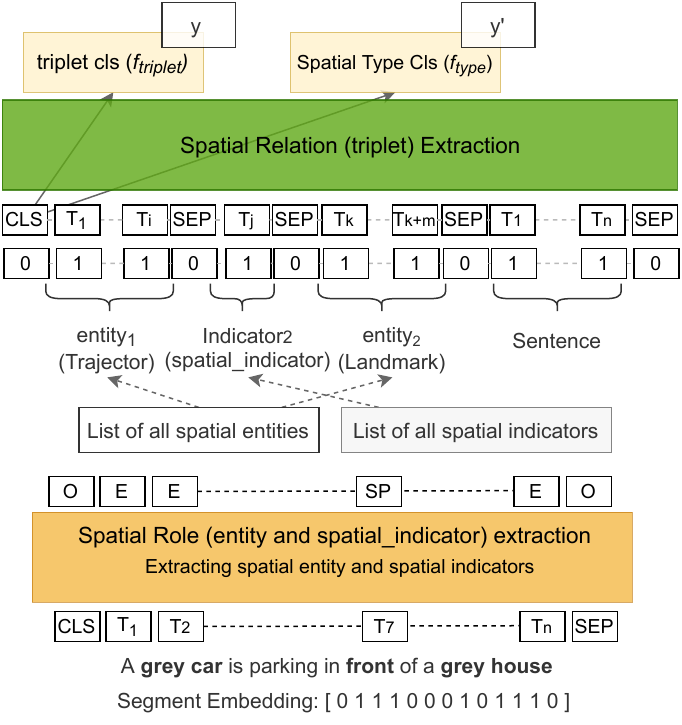}
    \caption{Spatial role labeling model includes two separately trained modules. E: entity, SP: spatial\_indicators. As an example, triplet (a grey house,  front , A grey car) is correct and the ``spatial\_type = FRONT'', and (A grey car,  front, a grey house) is incorrect, and the ``spatial\_type = NaN''. Image from~\cite{spartun}
    }
    \label{fig:pipeline}
\end{figure}

\subsection{Spatial Role Labeling~(\sprl{})}
The \sprl{} module, shown in Figure~\ref{fig:pipeline} is divided into three sub-modules, namely, spatial role extraction~(SRole), spatial relation extraction~(SRel)\footnote{Since the questions(Q) and stories(S) have different annotations~(questions have missing roles), we separately train and test the SRel and SType modules}, and spatial type classification~(SType). We only use these modules on sentences that convey spatial information in each benchmark.
To measure the performance of \sprl{} modules, we use the macro average of F1 measure for each label. These modules are evaluated on three datasets that provide \sprl{} annotations, \msprl{}, \spartqa{}, and \spartun{}. When training the \sprl{} module on auto-generated benchmarks, we achieved a performance of 100\%  using only a quarter of the training data, therefore we stopped further training. 


As shown in Table~\ref{tab:sprl-result}, all \sprl{} sub-modules achieve a high performance on synthetic datasets, \spartqa{} and \spartun{}.  This good performance is because
these datasets may contain less ambiguity in the natural language expressions. Therefore, the \bert{}-base models can easily capture the syntactic patterns needed for extracting the roles and direct relations from the large training set. 

\subsection{Coreference Resolution~(Coref) in Spatial Reasoning}
\label{appendix:coref}

\begin{figure}[h]
	\centering
	\begin{subfigure}[t]{\linewidth}
	
		\includegraphics[width=\linewidth]{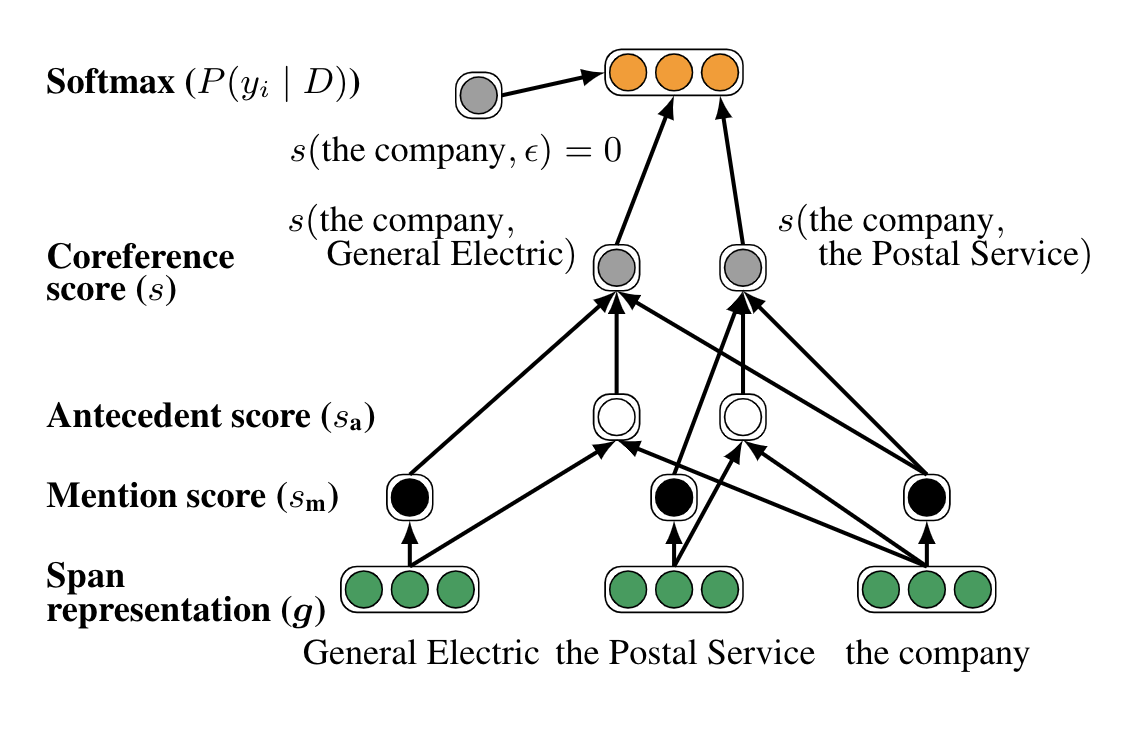}
		\caption{The coreference resolution model structure.}
		\label{fig:coref}
	\end{subfigure}
	\begin{subfigure}[t]{\linewidth}
		\includegraphics[width=\linewidth]{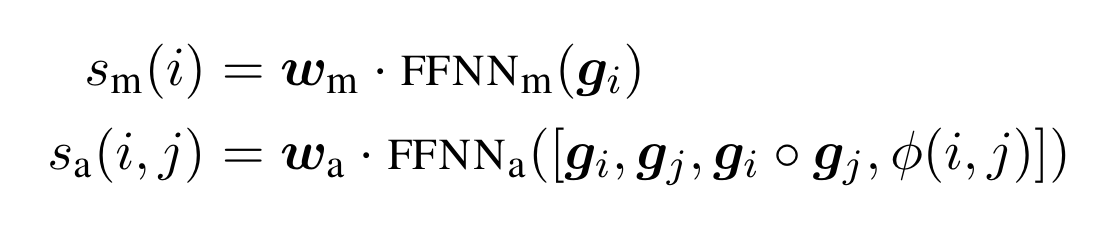}
		\caption{The formula for computing the coreference scores}
		\label{fig:formula}
    \end{subfigure}
	\hfill
 	\caption{The coreference resolution model~\cite{lee-etal-2017-end}.}
	\label{fig:coref-w}
\end{figure}



	

 We implement a coreference resolution model based on~\cite{lee-etal-2017-end} to extract all antecedents for each entity~(check Figure~\ref{fig:coref}).
Compared to the previous works, we have the entities~(phrase) annotations. Hence, we ignore the phrase scoring modules and use this annotation instead. 
 We first collect all mentions of each predicted entity from spatial role extraction or role annotations, then assign an ``id'' to the same mentions and include that id in each triplet. For example, for BELOW(a cat, a grey car), Front(the car, a church), id 1= {a cat}, 2 = {a grey car, the car}, and 3 = {a church}. So we create new triplets in the form of BELOW(1, 2) and Front(2, 3).
 
To train the model, we pair each mention with its previous antecedent and use cross-entropy loss to penalize the model if the correct pair is not chosen. For singletons and starting mention of objects, the model should return class $0$, which is the $[CLS]$ token. Since the previous model does not support the plural antecedent~(e.g., two circles), we
include that by considering shared entity in pairs like both (two circles, the black circle) and (two circles, the blue circle) are true pairs.

As an instance of the importance of coreference resolution in spatial reasoning, consider this context ``block A has one black and one green circle. The black circle is above a yellow square. The yellow square is to the right of the green circle. Which object in block A is to the left of a yellow square'' The reasoner must know that the NTPPI(block A, one green circle) and RIGHT( the yellow square, the green circle) are talking about the same object to connect them via transitivity and find the answer.

\textbf{To evaluate} the coreference resolution module~(Coref in Table~\ref{tab:sprl-result}), we compute the accuracy of the pairs predicted as Corefs. 
The Coref model achieves a high performance on all datasets. 
The performance is slightly lower on the \human{} dataset when \spartun{} is employed for additional pre-training. However, we observed that there are many errors in the annotations in \human{}, and the pre-trained model is, in fact, making more accurate predictions than what is reflected in the evaluation.


\subsection{Logic-based Spatial Reasoner}
\label{appendix:rules}

\begin{table}[t]
\resizebox{\linewidth}{!}{%
\begin{tabular}{|l|c|c|ccc|}
\hline
\textbf{Datasets}           & \textbf{\qtype{}}      & \textbf{Total} & \textbf{A} & \textbf{C} & \textbf{R} \\ \hline
\multirow{2}{*}{\auto{}} &YN & 18             & 7           & 10          & 1           \\
    & FR & 38             & 5           & 20          & 13          \\ \hline

\multirow{2}{*}{\spartun{}} & YN & 13            & 4           & 9           & 0           \\
    & FR & 35             & 0           & 35          & 0           \\ \hline
\multirow{1}{*}{\human{}} &YN &  29            &   20          &    6       &   3         \\
        \hline
\end{tabular}%
}
\caption{Analyzing wrong predictions in GT-\etor{}. A:  Missing/errors in \textbf{A}nnotation, C: rule-based \textbf{C}oreference issues in connecting extracted information, R: Shortcomings of the \textbf{R}easoner.}
\label{tab:noise_report}
\vspace{-3mm}
\end{table}

\textbf{To solely evaluate} the performance of the logic-based reasoner, we use the ``GT-\etor{}''.
We look into the errors of this model and categorize them based on the source of errors. The categories are  \textit{missing/wrong ground truth direct annotations}~(A), \textit{rule-based Coreference Error}~(C) in connecting the extracted information before passing to the reasoner, and \textit{the low coverage of spatial concepts in the reasoner}~(R).
As is shown in Table~\ref{tab:noise_report}, spatial Reasoner causes no errors for \spartun{} since the same reasoner has been used to generate it. 
However, the reasoner does not cover spatial properties of entities~(e.g., right edge in ``touching right edge'') in \spartqa{} and causes wrong predictions in those cases.